\documentclass[journal]{IEEEtran}

\ifCLASSINFOpdf

\else

\fi
\usepackage[english]{babel}
\usepackage[utf8x]{inputenc}
\usepackage{amsmath}
\usepackage{amsthm}
\usepackage{graphicx}
\usepackage[colorinlistoftodos]{todonotes}
\usepackage{mathtools}
\usepackage{subcaption}
\usepackage{amssymb}
\usepackage[export]{adjustbox}
\usepackage{stfloats}
\usepackage{lettrine}
\usepackage{tikz}
\usetikzlibrary{calc}
\newtheorem{remark}{Remark}

\begin{document}
\title{Over-the-Air Federated Learning in Satellite systems}

\author{Edward Akito Carlos,~\IEEEmembership{Fellow,~IEEE,} Raphael Pinard,~\IEEEmembership{Student Member,~IEEE}
        , Mitra Hassani}

%

% The paper headers
\markboth{}%
{Shell \MakeLowercase{\textit{et al.}}: Bare Demo of IEEEtran.cls for IEEE Journals}

\maketitle

% As a general rule, do not put math, special symbols or citations
% in the abstract or keywords.
\begin{abstract}
Federated learning in satellites offers several advantages. Firstly, it ensures data privacy and security, as sensitive data remains on the satellites and is not transmitted to a central location. This is particularly important when dealing with sensitive or classified information. Secondly, federated learning allows satellites to collectively learn from a diverse set of data sources, benefiting from the distributed knowledge across the satellite network. Lastly, the use of federated learning reduces the communication bandwidth requirements between satellites and the central server, as only model updates are exchanged instead of raw data. By leveraging federated learning, satellites can collaborate and continuously improve their machine learning models while preserving data privacy and minimizing communication overhead. This enables the development of more intelligent and efficient satellite systems for various applications, such as Earth observation, weather forecasting, and space exploration.
\end{abstract}

% Note that keywords are not normally used for peerreview papers.

\IEEEpeerreviewmaketitle

\section{Introduction}
Federated learning in the context of satellites refers to a distributed machine learning approach where multiple satellites collaboratively train a model without sharing their sensitive data with a central server or each other.

In this scenario, each satellite processes and learns from its own locally collected data, using onboard computational resources. The satellites exchange model updates rather than raw data, leveraging inter-satellite communication links or ground stations. The updates are aggregated in a centralized manner or through a decentralized scheme, allowing the collective intelligence of the satellite network to improve the global model.

Federated learning in satellites offers several advantages. Firstly, it enables efficient utilization of computational resources on individual satellites, reducing the need for extensive data transmission and storage. This is particularly beneficial in scenarios where bandwidth and storage capacity are limited. Additionally, federated learning ensures data privacy and security by avoiding the need to transmit sensitive information between satellites or to a central server.

Moreover, the distributed nature of federated learning in satellites enhances robustness and resilience. If a satellite encounters communication disruptions or malfunctions, other satellites can continue training the model independently. This decentralized approach also supports scalability, allowing new satellites to join the federation without requiring a centralized retraining process.

By leveraging federated learning, satellite networks can collectively improve their machine learning models while maintaining data privacy, optimizing resource usage, and enhancing overall performance and adaptability in space-based applications.

Federated learning in the context of satellites involves the application of collaborative machine learning techniques within a network of satellites. It allows satellites to collectively train machine learning models while preserving data privacy and minimizing communication overhead.

In satellite systems, federated learning can be employed to leverage the data collected by individual satellites without the need to transmit sensitive or large-scale data back to a central server on Earth. Instead, each satellite performs local model training using its onboard data and computational resources.

The trained models are then shared among the satellites, either directly or through relay satellites, using inter-satellite communication links. These models can be combined or averaged to create an improved global model that captures knowledge from each participating satellite.

Federated learning in satellites offers several advantages. Firstly, it enables collaborative learning across a distributed network of satellites, allowing the utilization of a larger and more diverse dataset. Secondly, it reduces the need for extensive data transmission to Earth, which can be costly and inefficient. Moreover, it addresses privacy concerns by keeping sensitive data onboard the satellites and sharing only aggregated model updates.

By employing federated learning techniques, satellite systems can benefit from collective intelligence and improved models while optimizing communication bandwidth and preserving data privacy in a distributed environment.

\section{Satellite arrangement to achieve maximum channel capacity} \label{sec:max}
In this section, we aim to find the optimum satellites' position which maximizes the channel capacity. For this, we introduce a new variable $\nu_i=\sin(\theta_i)\sin(\phi_i)$. Also, we assume that $d\geq \frac{\lambda}{2}$, and therefore $kd\geq \pi$
\\
From Arithmetic Mean-Geometric Mean inequality we obtain
\begin{align}  \label{eq:45}
C= \sum_{i=1}^{n_T}  \log_2 \big( 1+\frac{P}{\sigma^2} \lambda_i \big)= \log_2 \Big( \prod_{i=1}^{n_T}\big( 1+\frac{P}{\sigma^2} \lambda_i \big)  \Big) \nonumber \\
\leq  \log_2 \Big(\sum_{i=1}^{n_T} \big( 1+\frac{P}{\sigma^2} \lambda_i \big) \Big) ~~~ ~~~
\end{align}
On the other hand, because $\sum_{i=1}^{n_T} \lambda_i= Trace(\mathbf{W})=1$, inequality  \eqref{eq:45} becomes 
\begin{align}  \label{eq:46}
C \leq  \log_2 \Big( n_T+\frac{P}{\sigma^2} \Big) ~~~ ~~~
\end{align}
and equality happens when all $\lambda_i$'s are equal. Also, we know that all diagonal elements of matrix $\mathbf{W}$ are equal, meaning that if the off-diagonal elements become zero, we reach the maximum capacity. 

We divide the problem into three cases:
\begin{itemize}
\item{$n_T=n_R$}
In this case $\mathbf{W}=H^\dag H$. If $H^\dag $, or equivalently $H$, is unitary, then $\mathbf{W}$ would be a diagonal with them same eigenvalues. If $(kd \nu_i)$'s are evenly distributed on the unit complex circle, H becomes unitary; because in \eqref{eq:17} off diagonal elements become zero. Hence:
\begin{align}  \label{e:47}
kd\nu_i=\frac{2\pi \times i}{n_T}+ \mu_0 ~~ i=1,2,...,n_T
\end{align}
where $ -\frac{2\pi}{n_T}-\pi \leq \mu_0 \leq -\pi$ and it is such that $-\pi \leq kd\nu_i \leq \pi$.
\begin{remark}
Because $kd\geq \pi$,  $ -1\leq \nu_i=\sin(\theta_i)\sin(\phi_i) \leq 1$, which guarantee that $\theta_i$ and $\phi_i$ have solutions. 
\end{remark}
\item{$n_T<n_R$}
Again $\mathbf{W}=H^\dag H$. In this case if $H^\dag $, or equivalently $H$, is semi-unitary then $\mathbf{W}$ is diagonal. To satisfy this, $n_T$ vectors of length $n_R$ must be mutually orthogonal over Hermitian inner product space. 
\\
First we define the following set with $n_R$ elements
\begin{align}  \label{eq:48}
kd \tilde{\nu_i}=\frac{2\pi \times i}{n_R}+ \mu_0 ~~ i=1,2,...,n_R
\end{align}
where $ -\frac{2\pi}{n_R}-\pi \leq \mu_0 \leq -\pi$ and is constant. Each subset of the above set with cardinality of $n_T$ makes the matrix H' semi unitary.  
\item{$n_T>n_R$}
\\
In this case,  $\mathbf{W}=H H^\dag $, and hence the same discussion as case 2 could be held.
\end{itemize}
In the following figure, one realization of the satellite arrangement to achieve maximum capacity when   $n_T=n_R=4$ is depicted.
\\
To satisfy the above conditions, for instance, we begin with $\sin(\theta_i)=0.9$ for all users which means $\theta_i=0.35\pi$ for $i=1,2,3$ and $4$. This yields $\sin(\phi_1)=\frac{-5}{6}$, $\sin(\phi_2)=\frac{-5}{18}$,  $\sin(\phi_3)=\frac{5}{18}$ and  $\sin(\phi_4)=\frac{5}{6}$, or $(\phi_1,\phi_2,\phi_3,\phi_4)=(-0.69\pi,-0.08 \pi, 0.69\pi, 0.08 \pi)$

\begin{remark}
We found values for $\nu_i$'s which maximize the channel capacity in the above cases. When linear antenna is along x-axis $\nu_i=\cos(\theta_i)\sin(\phi_i)$ and when it is along the z-axis it is $\nu_i=\cos(\theta_i)$. Thus, the values found for $\nu_i$'s could be used correspondingly to maximize the channel capacity based on the antenna alignment. 
\end{remark}

\begin{remark}
$\nu_i$'s are not unique, because $\mu_0$ is an interval. Once the value for $\mu_0$ is chosen, it would be fixed for all $\nu_i$'s.
\end{remark}

\begin{remark}
Even for a constant $\mu_0$, for the antenna align X or Y axis, satellite configuration is not unique. ;However, the configuration would be unique for an antenna along Z axis. 
\end{remark}

\begin{figure}[htbp]
\centering{\includegraphics[scale=0.4]{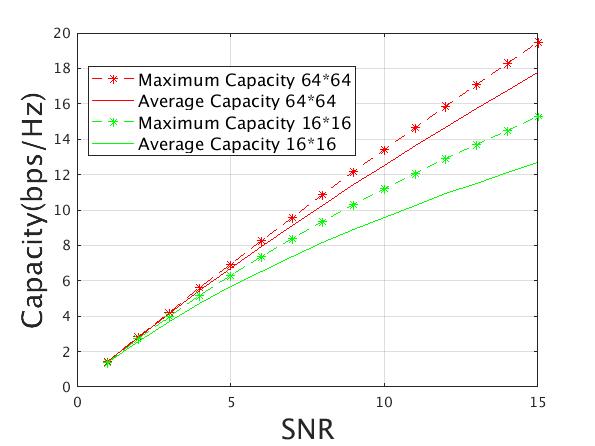}}
   \caption{Maximum Vs. Average channel capacity.}
   \label{fig:max}
\end{figure}

In Figure \ref{fig:max}, the maximum capacity is compared with average capacity for 2 cases.

\section{arrays with other configurations} \label{sec:config}
This section considers two other possibilities for the antenna configuration which are mentioned in the following parts
\subsection{Linear array not along y axis}
One can derive the results we obtained so far for any other alignment of linear array antenna. This could be done by finding by rotating the array to make it along y-axis and then finding the corresponding $\phi$ and $\theta$ values. 
\\
For an array placed on z axis, the array factor in \eqref{eq:4} becomes
\begin{align} \label{eq:49}
AF(\theta, \phi)=\sum_{m=0}^{M-1} I_m e^{jkmd\cos(\theta)}
\end{align}
Only $\theta$ appears in matrix $H$ which makes the calculation simpler, and all the results can be obtained similarly for this array. In appendix D we have found, for instance, $\mathbf{E} \{C^2 \}$ for this alignment.
\subsection{Rectangular array}
The elements are arranged uniformly along a rectangular grid in the xy plane to form an $m \times n$ array. This array could be regarded as either $m$ or $n$ linear array with $n$ and $m$ antenna elements, respectively. Henceforth, each sub-array can be analyzed independently, and the result would be added up. 

\section{minimizing channel interference} \label{Sec:Kissing}
In this section, we assume that we want to put some satellites around the Earth serving terrestrial users such that the amount of interference is minimized. To do so, firstly, we put satellites so that the minimum distance between them (considering all possible pairs) is maximized. This question intuitively reminds us of the “Tammes problem” which has been extensively studied before. \\
Tammes \cite{tarnai1991covering} problem looks for an answer for the following question: “How must N congruent non-overlapping spherical caps be packed on the surface of a unit sphere so that the angular diameter of spherical caps will be as great as possible”\\
One can easily correspond our problem to the answer of Tammes problem. We
can say all satellites are located on a unique sphere when revolving around the
Earth. It is worth mentioning that when a satellite is relatively close
to Earth, the orbit on which the satellite traverses is roughly a circle, and therefore our assumption is valid.\\
Tammes problem was solved for some specific number of points(for N=1,2,…,12,23,24 and some other values \cite{schutte1951kugel} \cite{van1952punkte} \cite{robinson1961arrangement}).\\

Let X be a finite subset of $S^{n-1}$ in $\mathbb{R}^{n}$. We define $\psi$ as follows:
\begin{align}
\psi(x)=\min_{x,y \in X} {dist(x,y)},   x\neq y
\end{align}
Then X is a spherical $\psi(X)$-code. Also, define $d_{N}$ the largest angular separation $\psi(X)$ with $|X|=N$ that could be obtained in $S^2$, meaning that:
\begin{align}
d_{N}=\max_{X \subset S^2} {\psi (X)}, |X|=N.
\end{align}

The remaining of this section considers two cases, one when there is no interference and not all parts of Earth is served by satellites, one when interference exists and all Earth surface is served. Afterward, channel capacity is calculated for the setup with minimum interference and it is compared with that of obtained in \ref{sec:max}.

\subsection{No interference}
In this part, we assume that each non-overlapping spherical cap found for each N is the terrestrial coverage area for the corresponding satellite. In this case, each server on Earth is served by a single satellite, and therefore there is no interference. Having this setup, there exist some place on Earth not being served by any satellites. For this matter, we define coverage percentage for each configuration as ratio of the area covered by satellites to surface area of Earth. 
\\
The following table could be accordingly attained.  
\\
As seen, the coverage percentage is non-linear as N grows, however, it reaches it maximum value for N=12 among the values considered above. 
\begin{table}[t]
 \centering
 \begin{tabular}{|l|c|c|c|c|c|}
 \hline
 Number of satellites & $d_{N}$ & Coverage percentage\\
 \hline
4   &  109.4712206 &  0.8386\\
5  &  90.0000000&  0.7322\\
6  &  90.0000000 &  0.8787\\
7  & 77.8695421 & 0.7775\\
8  & 74.8584922 & 0.8234\\
9  & 70.5287794 & 0.8258\\
10  & 66.1468220 & 0.8101 \\
11  & 63.4349488 & 0.8214\\
12  & 63.4349488 & 0.8961 \\
13  & 57.1367031 & 0.7914\\
14  & 55.6705700 & 0.8099\\
15 & 53.6578501 & 0.8073\\
16 & 52.2443957 & 0.8171\\
17 & 51.0903285 & 0.8309\\
\hline
\end{tabular}
\caption{No interference setup. $d_{N}$ and coverage percentage is shown for different number of satellites around the Earth}
\label{tab1}
\end{table}
\subsection{Interference and Overlapping Coverage Area} \label{sub:inter}
to cover the whole surface of Earth with existing satellites, the coverage area for each satellite needs to be enlarged. For this purpose, each spherical cap is equally enlarged till all points on the surface of the Earth would be covered by at least on satellite.
\\ 
In this case, there would be some terrestrial servers receiving signals from a couple of satellites. From \cite{mooers1994tammes}, for $N>6$, if a server receives signal from more than 1 satellites, the number of satellites seen by the server is at most 5 and at least 3. 
\\
In \cite{tarnai1991covering}, the author tries to find conjectured solutions for this problem. Also, it defined density denoted by $D_{N}$ which has the same meaning as Coverage Percentage defined in this paper. 

\begin{table}[t]
 \centering
 \begin{tabular}{|l|c|c|c|c|c|}
 \hline
 Number of satellites & $d_{N}$ & Coverage percentage\\
 \hline
4   & 70.5287 & 1.3333\\
5  &  63.4349 & 1.3819\\
6  &  54.7356 & 1.2679\\
7  & 51.0265 & 1.2986\\
8  & 48.1395 & 1.3307\\
9  & 45.8788 & 1.3672\\
10  & 42.3078 & 1.3023\\
11  & 41.4271 & 1.3761\\
12  & 37.3773 & 1.2320\\
13  & 37.0685 & 1.3135\\
14  & 34.9379 & 1.2615\\
15 & 34.0399 & 1.2851\\ 
16 & 32.8988 & 1.2829\\
17 & 32.0929 & 1.2989\\
\hline
\end{tabular}
\caption{Serving all parts of the Earth with $N$ satellites.  Coverage percentage $>1$. $d_{N}$ and coverage percentage are shown for different number of satellites around the Earth.}
\label{tab2}
\end{table}

\subsection{Channel capacity}
To compare the channel capacity for the satellite arrangement in\ref{sub:inter} to that of found in \ref{sec:5}, we only considers the satellites located in one hemisphere (The one that the receiver is located). We used the method elaborated in \cite{tarnai1991covering} to find the satellites location in the desired hemisphere. 
\begin{figure}[htbp]
\centering{\includegraphics[scale=0.4]{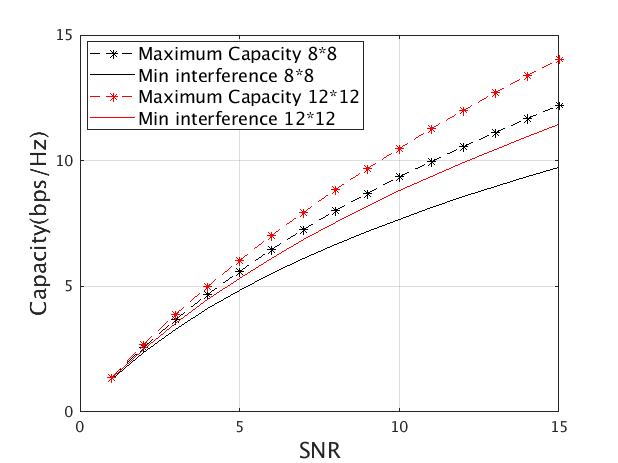}}
   \caption{Maximum channel capacity Vs. channel capacity for the setup with minimum interference.}
   \label{CDD}
\end{figure}

\section{Rotman lens} \label{sec:Rotman}
\begin{figure}[htbp]
\centering{\includegraphics[scale=0.3]{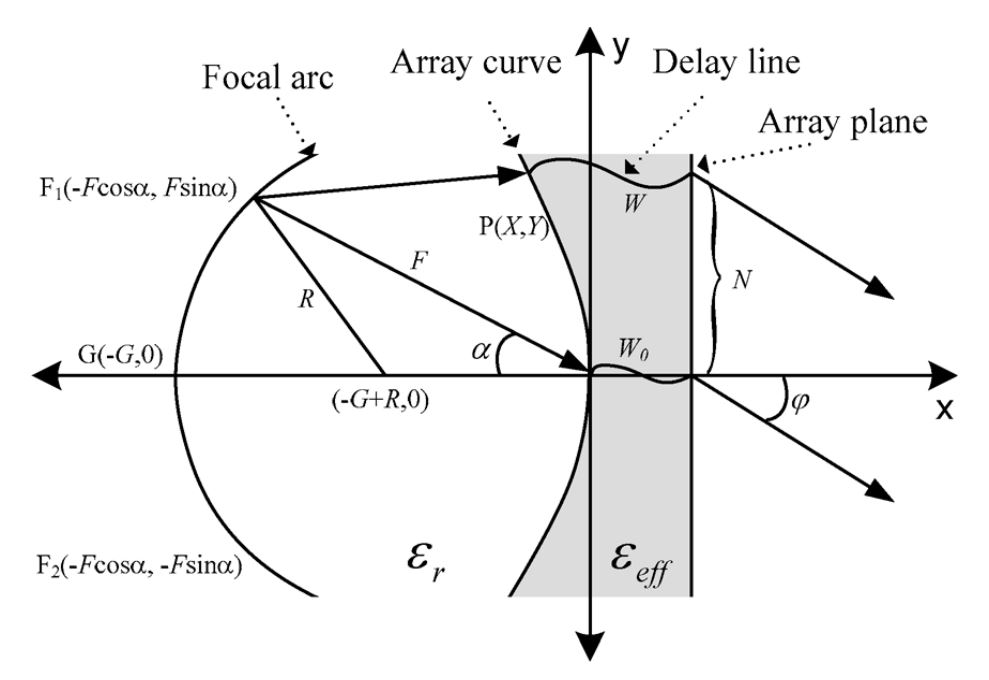}}
   \caption{Microwave Rotman Lens geometry and design parameters \cite{lee2009beamforming}.}
   \label{Rotman}
\end{figure}

This lens is capable of beam-forming without the need for switches or phase shifters. Its typical geometry and design parameters are shown in Figure \ref{Rotman}.

Since the invention of Rotman lens, the lens equation has been extensively studied \cite{kim2001scaling} , \cite{katagi1984improved,hamidi2019systems,hamidi2022over,struhsaker2020methods}. Specifically, for a KU-band Rotman lens, the normalized transfer matrix model obtained in \cite{rahimian2013design} as follows
\\
\\
\begin{equation}
\begin{aligned}  \label{eq:S}
\mathbf{S}=\frac{e^{jkW_{0}}}{\sqrt{n_Tn_R}}~~~~~~~~~~~~~~~~~~~~~~~~~~~~~~~~~~~~~~~~~~~~~~~
\\
\times
\begin{pmatrix*}[c]
e^{-jk\eta_{1}\sin(\theta_1)} &. . .&  e^{-jk\eta_{1}\sin(\theta_{n_T})}\\
e^{-jk\eta_{2}\sin(\theta_1)} & . . .&  e^{-jk\eta_{2}\sin(\theta_{n_T})}\\
.& ...& .\\
.& ...& .\\
.& ...& .\\
e^{-jk\eta_{n_R}\sin(\theta_1)} & . . .& e^{-jk \eta_{n_R}\sin(\theta_{n_T})}\\
\end{pmatrix*}~~~~~~~~~~
\\
\times
\begin{pmatrix*}[c]
e^{-jk(F_1+V_1)} &. . .& 0 & 0\\
0 & . . .&  e^{-jk(F_m+V_m)}& 0\\
.& ...& .\\
.& ...& .\\
.& ...& .\\
0 & . . .& 0 & e^{-jk(F_{n_T}+V_{n_T})} \\
\end{pmatrix*}
\end{aligned}  
\end{equation}
\\
\\
where $\eta_{n}=[n-\frac{(N+1)}{2}]\times d$ and $e^{-jk(F_m+V_m)}$ is an extra phase shift added to the $m^{th}$ RF beam port; and $(n=1,2,...,n_T), (m=1,2,...,n_R)$. 
The transfer matrix has two parts: the common phase-delay shared by the array ports and the progressed phase through array port in which the latter contributes to RF beam forming.
phase alignment requires the phase-delay path (i.e., $F_{m}+V_{m}$) to be constant so that the Rotman lens retain the true-time-delay (TTD) characteristic of the device. This method is essentially required in the case of simultaneous multiple RF beam port excitation to obtain reconfigurable far-field microwave radiation properties for the satellite RF lens \cite{zhang2012reconfigurable}, Meaning that
\begin{align}  
e^{-jk(F_m+V_m)}=constant
\end{align}
and therefore, the second matrix in \eqref{eq:S} becomes identity matrix. 
\\
If the beam angles $\theta_m$'s received from $n_T$ satellites are uniformly distributed over $(0, \frac{\pi}{2})$ and there are $n_R$ antenna elements over the array plan, the average channel capacity  of this setup could be obtained similar to that of obtained in Appendix \ref{Z axis}, i.e. a linear array along Z axis.
\section{conclusion} \label{sec:conclusion}
This paper analyzes the channel capacity of a MIMO system in which the receiver is equipped with a linear array antenna and the transmitters are some satellites whose locations are not known at the receiver. To address the characteristic of this MIMO channel, the average channel capacity is found and also the outage probability is presented. In addition, the optimum position of satellites are found such that the channel capacity is maximized. In contrast, the optimum position of satellites to cause minimum interference at the receiver is presented and the channel capacity in this setup is compared with the maximum possible channel capacity. As an application of this paper, the average capacity for a Rotman lens performing for KU-band satellites is also analyzed.

\bibliographystyle{IEEEtran}
\bibliography{refs}
\end{document}